\documentclass[a4paper]{article}

\usepackage{arxiv}

\usepackage[utf8]{inputenc}
\usepackage[T1]{fontenc}
\usepackage{url}
\usepackage{graphicx}
\usepackage{natbib}
\usepackage{amsmath}
\usepackage{algorithm}
\usepackage{algpseudocode}

\title{Extrapolated Markov Chain Oversampling Method for Imbalanced Text Classification\thanks{The authors acknowledge support from the Academy of Finland via the Finnish Centre of Excellence in Randomness and Structures (decision number 346308). Moreover, Aleksi Avela acknowledges the personal grants from The Emil Aaltonen Foundation (Nuoren tutkijan apuraha, numero 230014, and työskentelyapuraha, numero 240013).}}

\author{
 Aleksi Avela \\
  Aalto University, School of Science\\
  Department of Mathematics and Systems Analysis\\
  \texttt{aleksi.avela(at)aalto.fi} \\
   \And
 Pauliina Ilmonen \\
  Aalto University, School of Science\\
  Department of Mathematics and Systems Analysis\\
  \texttt{pauliina.ilmonen(at)aalto.fi} \\
}

\begin{document}
\maketitle
\begin{abstract}
Text classification is the task of automatically assigning text documents correct labels from a predefined set of categories. In real-life (text) classification tasks, observations and misclassification costs are often unevenly distributed between the classes---known as the problem of imbalanced data. Synthetic oversampling is a popular approach to imbalanced classification. The idea is to generate synthetic observations in the minority class to balance the classes in the training set. Many general-purpose oversampling methods can be applied to text data; however, imbalanced text data poses a number of distinctive difficulties that stem from the unique nature of text compared to other domains. One such factor is that when the sample size of text increases, the sample vocabulary (i.e., feature space) is likely to grow as well. We introduce a novel Markov chain based text oversampling method. The transition probabilities are estimated from the minority class but also partly from the majority class, thus allowing the minority feature space to expand in oversampling. We evaluate our approach against prominent oversampling methods and show that our approach is able to produce highly competitive results against the other methods in several real data examples, especially when the imbalance is severe. 
\end{abstract}


\vspace{5mm}

\section{Introduction}
\label{sec:intro}

In the field of machine learning, natural language processing is of uttermost and ever-increasing importance. Text classification (or text categorization) is the task of assigning natural language text documents (e.g., emails, news articles, or abstracts of scientific papers) correct labels from a predefined set of categories based on a set of automated decision rules. In this work, with text classification we refer to supervised machine learning text classification.

Document embedded approaches are commonly applied in text classification tasks. Perhaps the most used document level approach is the bag-of-words model, where each index $i$ in a document vector accounts for the count of word $w_i$ in the given document. Documents may also be embedded into vectors, for example, based on character or word $n$-grams as well. The bag-of-words model is a well-studied approach and, as the observations are modeled with fixed-sized vectors, it allows applying any standard classification algorithms, such as support vector machines or naive Bayes classifiers, see, e.g., \cite{joachims02} and \cite{rennie03}.

In practical text classification applications, for instance, in information retrieval, spam filtering, and fraud detection, it may be that the considered classes are not balanced. That is, often one or some of the classes are notably larger than the other(s) and, on top of that, observations in the rare class(es) commonly have higher misclassification costs compared to the common observations. The set of issues related to uneven class and cost distributions is referred to as the problem of imbalanced data, see, e.g., \cite{japkowicz00}.

Imbalanced class and misclassification cost distributions are commonly caused by the fact that real-life data sets often consist, for the most part, of ``typical'' observations, and only a minor share of the data represents ``interesting'' observations \citep{chawla02,weiss04}. Multiclass data sets often cause high imbalance in classification, as, commonly, the approach to multiclass classification with $k$ classes is to train $k$ separate binary classifiers with a one-versus-rest (OvR) scheme. Especially when the number of classes $k$ is large, dividing classification into $k$ binary tasks increases imbalance in each classification task \citep{bishop06}. In this article, we consider a number of multiclass data sets with OvR approach. Each individual classification task consists of a positive (minority) class and a negative (majority) class which is composed of all the rest of the categories.

Standard classifiers often become inefficient when dealing with imbalanced data and may end up favoring the majority class over the minority class \citep{he09,japkowicz00,lopez12}. One of the main reasons for this is that most classification algorithms have been developed, either explicitly or implicitly, to maximize (in-sample) accuracy \citep{weiss07}. This is problematic when considering an imbalanced data set, as in practice, observations that belong to the minority class are often the ones that are desired to be detected, i.e., their misclassification costs are higher.

Accuracy is not a suitable statistic for evaluating classification performance on an imbalanced data set with uneven misclassification costs \citep{chawla02}. For instance, if a bank is considering whether to reject a written application for a loan, it is clear that a false negative (i.e., lending money to a person that will not repay the loan) has a higher cost than a false positive (i.e., declining an applicant that would have repaid their loan). There are other issues related to imbalanced data as well. \cite{jo04} argue that minority classes tend to be heterogeneous and to consist of small disjuncts, and therefore, generalizing the minority class is often difficult for the classifier.

Theoretically, an optimal (cost-sensitive) classifier minimizes the total conditional expected value of misclassification costs \citep{elkan01}. Building an optimal cost-sensitive classifier, however, is often difficult as many classification algorithms are poor at providing reliable posterior probability estimates \citep{iranmehr19,zadrozny01}. Moreover, assigning explicit misclassification costs may be challenging in real-life tasks \citep{weiss04,weiss07}. In addition to cost-sensitive learning, there exists two other main categories of approaches for classification of imbalanced data: sampling techniques, where the training data is modified to better suit the properties and strengths of machine learning classification algorithms, and algorithmic approaches, where the learning algorithm itself is modified to consider the uneven class distribution and misclassification costs, see, e.g., \cite{veropoulos99}.

In practice, sampling approaches are highly popular, mainly since they allow the use of any classification algorithm and do not require assigning explicit misclassification costs. However, it can be argued that applying sampling before training a classifier is implicitly posing some misclassification costs on the considered classes as well \citep{weiss07}. Resampling can be executed by undersampling the majority class, by oversampling the minority class, by hybrid methods that combine undersampling and oversampling, or by an ensemble of classifiers trained on multiple different resampled training sets that form the final classification as a majority vote. For an overview of (sampling and other) approaches to imbalanced classification, we refer the reader, for example, to \cite{haixiang17} and \cite{he09}, and the references therein.

In this article, we introduce a novel Markov chain based oversampling method for imbalanced text data. Differing from many general-purpose oversampling methods, in our approach, the synthetic observations are not bounded by the convex hull of the minority sample. Our method is designed particularly for classification of smaller text data sets with severe imbalances and for tasks where deep learning based approaches may not be an option due to, e.g., lack of computational resources, privacy issues, or limited training data. That is, our approach mainly focuses on remedying challenges related to document-embedded sampling and classification approaches to imbalanced text data. However, we also provide an example of how our method can produce competitive results with a word-embedded neural network approach to imbalanced text classification.

The rest of this article is organized as follows. In Section \ref{sec:background}, we provide an overview of oversampling in general, and in Section \ref{sec:oversampling}, we discuss oversampling text data. In Section \ref{sec:emco}, we first motivate our approach and then describe the method. Section \ref{sec:experiments} describes the experiments, Section \ref{sec:results} presents and discusses the obtained classification results, and Section \ref{sec:vocabulary} evaluates feature space growth dynamics of the applied text oversampling methods. Finally, Section \ref{sec:conclusion} concludes the work.

\section{Background}
\label{sec:background}

Oversampling is a popular approach for handling imbalanced data, and many studies have shown that it can greatly improve classification performance, see, e.g., \cite{chawla02}, \cite{chen11}, \cite{cloutier23}, \cite{das15}, \cite{deng24}, \cite{he08}, \cite{moreo16}, \cite{nakada24}, and \cite{weiss07}. Contrary to undersampling, oversampling does not waste any data, which is often a desired feature if one is working with limited amounts of labeled training data. On the other hand, unlike cost-sensitive learning and algorithmic approaches, oversampling can easily be combined with any classification algorithm.

The most direct way to oversample the minority class is to randomly duplicate existing minority observations (that is, sample with replacement) until the desired balance level is reached. This approach is referred to as random oversampling. However, random oversampling is known to be prone to cause overfitting as it may lead to the learning algorithm overestimating the significance of existing minority examples when fitting the decision rule(s).

In order to address the issue of overfitting, oversampling methods have often been designed to oversample the minority class by creating synthetic observations. Synthetic oversampling has been a more popular sampling approach than random oversampling ever since \cite{chawla02} introduced the idea. In this article, with oversampling, we refer to the process of creating synthetic observations, that is, observations that did not (necessarily) appear in the training set before sampling. The majority of oversampling methods are non-parametric as they do not require any assumptions of data generating distributions. Nevertheless, researchers have also presented probabilistic oversampling approaches, see, e.g., \cite{chen11} and \cite{das15}.

The Synthetic Minority Over-sampling Technique (SMOTE) developed by \cite{chawla02} and its variants are among the most applied oversampling approaches \citep{piyadasa23}. SMOTE is based on generating synthetic observations as random convex combinations of pairs of minority training examples. The pairs are formed by selecting one minority observation at a time and randomly choosing one of its $k$ nearest neighbors. The number of generated synthetic observations for each pair depends on the imbalance ratio and the desired class balance ratio after sampling.

Some of the most popular variants of SMOTE include, for instance, Geometric-SMOTE \citep{douzas19}, where synthetic observations are generated in a geometric region instead of a line connecting two examples, Borderline-SMOTE \citep{han05}, that generates synthetic observations only close to the region separating the classes, and Adaptive Synthetic (ADASYN) sampling \citep{he08}, which is a density-based variant of SMOTE. ADASYN creates a varying number of synthetic observations in the neighborhood of each minority observation depending on how large share of its all $k$ nearest neighbors belong to the majority class. The idea is to focus on the border areas of the minority sample that are more difficult to learn for the classifier \citep{he08}.

Synthetic oversampling can also be viewed to relate to the field of data augmentation. The idea of data augmentation in supervised learning is to enrich training data with artificial observations generated from the original training examples in order to achieve better generalization and to reduce overfitting. In image recognition, for instance, data augmentation is rather straightforward to implement, e.g., by rotating, mirroring, and scaling the training examples. However, in text analysis, where such trivial transformations do not exist, augmenting data is a much more difficult task \citep{bayer22,shorten21}. For a survey of text data augmentation, we refer the reader, for instance, to \cite{bayer22} and \cite{henning23}, and the references therein.

\section{Oversampling text data}
\label{sec:oversampling}

Most of the work in research of oversampling methods has been done on a general level and without considering any specific domain of data \citep{chen11}. However, natural language is clearly a distinct type of data that differs considerably from many other domains. First of all, the feature space in text data (that is, the training vocabulary, when text is modeled as bag-of-words vectors) is often very large and sparse. The document embedded approaches are, either, not able to capture the semantic distances between different features. Moreover, text is sequential data, but modeling text documents as simple bag-of-word vectors loses the information about the word order.

Although there are some obvious drawbacks with modeling text data in document embedded spaces, these approaches have been shown to work extremely well in many text classification tasks. Applying document embedded approaches also allows the use of any general-purpose oversampling method for text data, such as SMOTE or ADASYN, as the training examples are represented with fixed sized vectors. These methods have been widely used in various applications with great results.

There exist also oversampling methods designed specifically for text classification. \cite{chen11} introduce an approach for text oversampling using a probabilistic topic model based on Latent Dirichlet Allocation (LDA) \citep{blei03}. The generative model assumes that every document in the minority class follows the same topic distribution. \cite{chen11} present two oversampling techniques based on the topic model approach: DECOM (data re-sampling with probabilistic topic models) and DECODER (data re-sampling with probabilistic topic models after smoothing). After estimating the probabilistic topic model, DECOM uses it for generating synthetic observations. On the other hand, DECODER first smooths the training set by regenerating the original training documents before oversampling.

\cite{moreo16} present a general Latent Space Oversampling (LSO) framework using latent semantic distributional properties to generate synthetic documents. Based on their LSO framework, \cite{moreo16} propose Distributional Random Oversampling (DRO) method, where the original feature space is extended with a latent space generated by relying on the distributional hypothesis. The distributional hypothesis \citep{harris54} states that words with similar document distributions possess somewhat similar semantics. First, DRO estimates the probabilities for the latent features. The size of the latent space equals the size of the training set. The estimated probabilities are then used for sampling random latent vectors that are concatenated to the original data vectors. On top of oversampling, DRO transforms the existing training and test observations into the new space as well.

Text augmentation based oversampling approaches include a wide variety of techniques ranging from synonym replacement to deep learning based generation \citep{henning23}. For instance, \cite{luo19} introduce an approach based on sequence generative adversarial networks (SeqGAN) that is suitable for oversampling large text data sets. Yet, also seemingly very simple ideas have proven to be highly useful in imbalanced text classification; Easy Data Augmentation (EDA) by \cite{wei19} is an example of a straightforward yet very effective method. EDA augments texts by doing synonym replacements, random word insertions, random swaps, and random deletion. Modern generative large language models (LLMs) provide also an opportunity for text oversampling, see, e.g., \cite{cloutier23} and \cite{nakada24}.

Synthetic sampling can reduce the risk of overfitting, but may not be able to fix the problem completely---especially if every synthetic example is bounded by the convex hull of the minority training set. Moreover, studies have argued that synthetic oversampling approaches that generate synthetic observations uniformly within the convex hull of the minority sample may be prone to over generalize the minority class \citep{he09}. The risk of overfitting with oversampling is an even more notable issue in text classification where the feature space does not necessarily remain immutable when the sample size grows. Often, increasing the sample size of text also increases the feature space, see, Section \ref{sec:motivation}. Vocabulary that appears in the minority training documents is usually a small subset of the complete training vocabulary. Commonly, this is not a property of the minority class itself but rather a consequence of its relative size compared to the majority class.

\section{Extrapolated Markov Chain Oversampling}
\label{sec:emco}

In this section, we introduce a novel text oversampling approach referred to as Extrapolated Markov Chain Oversampling (EMCO) method. First, we present motivation for the approach by addressing issues mentioned in the previous section, and then we describe the method.

\subsection{Motivation for extrapolated text oversampling}
\label{sec:motivation}

Empirical studies show that the size of vocabulary can be modeled as a function of the text sample size. This phenomenon is known as Heaps’ law \citep{heaps78} in information retrieval, and, with a slightly different formulation, as Herdan’s law in linguistics \citep{herdan64} (see also \cite{egghe07}). Heaps’ law models the relationship between the size of a text sample and the number of distinct words appearing in the sample as $T = k A^\theta$, where $T$ is the size of the vocabulary, $A$ is the total number of words in the sample, and $k > 0$ and $0 < \theta < 1$ are constants (depending on the context and language of the text sample) \citep{egghe07,heaps78}.

Inspired by this, we argue that, when the minority class is oversampled in text data, the minority feature space should expand as well. We suggest that the expansion of the synthetic minority vocabulary could be executed by considering some words in the majority training documents that do not appear in the minority training documents.

\begin{figure}[ht]
\centering
\includegraphics[width=0.68\textwidth]{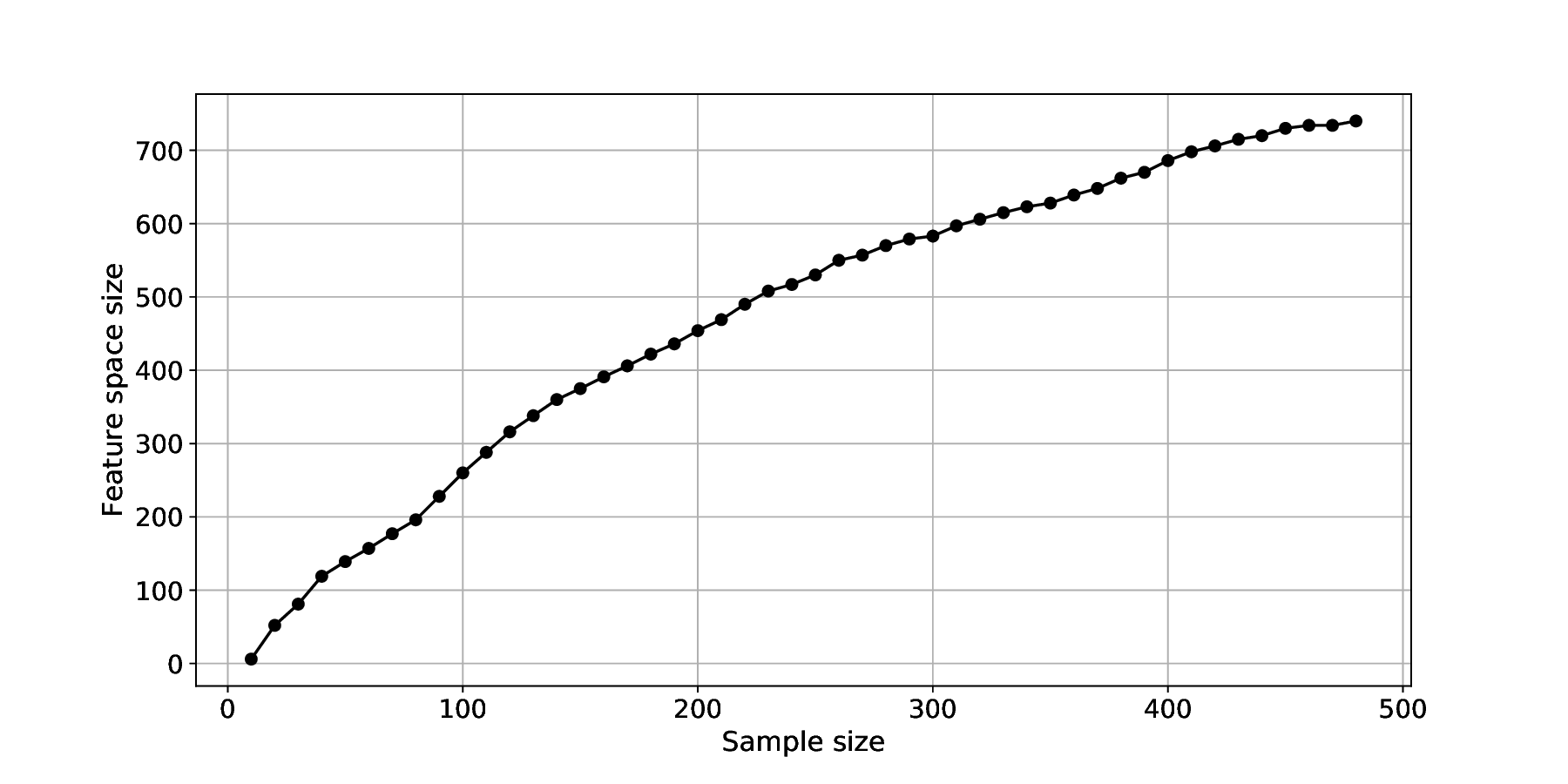}
\caption{The feature space of the ``trade'' category headlines in the Reuters-21578 data set expands when the sample size is increased. In each iteration, the sample size was increased by ten documents.}\label{Fig:featurespace}
\end{figure}

Feature space growth is illustrated in Figure \ref{Fig:featurespace}, where the size of the feature space is plotted over the sample size that is increased by including more documents in the sample. This figure was created using the headlines corresponding to one category (``trade'') of the news articles in the Reuters-21578 data set with stopwords removed.\footnote{The Reuters-21578 corpus is described in Section \ref{sec:experiments}.} When the sample size of the class ``trade'' increases from half to full size, the feature space grows by 211 words. Remarkable is that, in this particular example, 156 out of those words actually appear in the half-sized sample consisting of all the other documents (representing the majority class). That is, in this example, over 70\% of the new words are actually \emph{familiar} from the majority class, suggesting that some words in the majority documents could in fact be included in the synthetic vocabulary when oversampling the minority class. Note also that the pace of feature space growth seems to be faster the smaller the sample size is, emphasizing the importance of considering this property for small samples.

In this article, we base our approach on the idea that natural language is composed of (at least) two underlying structures: topic and sequence. Same idea plays a significant role in text generation, see, e.g., \cite{bowman15}, \cite{dieng16}, and \cite{wang19}, which is essentially a very similar task as text oversampling (albeit that, in our case, a document \emph{generated} by oversampling is not perceived as natural language but is transformed into the document embedded space).

We assume that, although each topic has its own marginal word probability distribution, regardless of the topic, certain words are still generally more likely to follow given words due to the natural sequential structure of text. For instance, let us assume that the topic of the considered document is sports news. Now, it could be argued that, due to the sequential structure, word ``high'' is more likely to be followed by word ``school'' than, e.g., by words ``match'' or ``scored'', even though the latter ones may be generally more likely to appear in this topic.

Modeling text documents as bag-of-words vectors is limited to the word distribution. We propose that, irrespective of the text modeling approach, the sequential structure of text could be considered in oversampling, too. Assuming that the structures of topic and sequence are, at least, partly independent in text data, some parts of the sequential information from the majority class could also be considered when oversampling the minority class.

We divide the training vocabulary into two mutually exclusive and collectively exhaustive word sets which we call \emph{minority vocabulary} and \emph{majority-only vocabulary}. The first set includes every distinct word that appears in the minority class training documents, and the second set consists of all distinct words that occur only in the majority training documents. That is, the \emph{minority vocabulary} may include a number of common words that appear also in the majority training documents. In the majority documents, these common words can naturally be followed by words that belong to the \emph{majority-only vocabulary}.

Keeping in mind that our objective is to find a way to allow the minority feature space to expand in oversampling, we are particularly interested in the common words, i.e., those words that appear in both classes in the training data. By using the sequential structure of text, the common words could be a route to access the \emph{majority-only vocabulary} in oversampling. We make an assumption that the sequential structure of text is sufficiently independent of the topic so that those \emph{majority-only vocabulary} words that follow \emph{minority vocabulary} words in the majority training documents would be reasonable additions to the synthetic minority vocabulary.

The proposed EMCO approach is based on modeling the documents with a finite discrete-time Markov chain model, where the states represent the words in the training vocabulary. The idea of applying Markov chains for generating text is very natural and has been done many times before, see, for instance, \cite{abdelwahab18}, but our method introduces a new addition to a plain Markov chain model.

Other similar ideas have been applied to text classification earlier as well. For instance, \cite{nigam00} use unlabeled data together with labeled data for improving text classification performance. On the other hand, our method uses information in the majority class together with the minority class to improve oversampling. Both of these approaches are based on the idea that when we are dealing with data that has some specific structure (such as natural language), some information that is not directly related to the task (be it unlabeled data in classification or the majority class in oversampling) can be utilized as well by using the structure of the considered data.

We estimate the transition probabilities between different words from the training data. However, instead of just counting word transitions in the minority training set, we also consider certain pairs of consecutive words in the majority training documents. That is, subsequent words $w_i$ and $w_j$ in a majority class document are included in the estimation if, and only if, word $w_i$ belongs to the \emph{minority vocabulary}. However, note that word $w_j$ may be part of either \emph{minority} or \emph{majority-only vocabulary}.

The consequence of the described idea is that, when the estimated transition matrix is used for sampling synthetic documents as realizations of the Markov chain model, there are also non-zero transition probabilities to words that do not appear in the minority class training documents, and hence, we call the method \emph{Extrapolated} Markov Chain Oversampling. That is, oversampling with EMCO allows the minority feature space to grow, and thus, hopefully, mitigates the risk of overfitting when training the classifier.

\subsection{Estimation of EMCO}
\label{sec:estimation}

Markov chain (of order 1) is a stochastic sequence of random variables $X_1, X_2, X_3, ... $ that has the Markov property, that is:

\begin{align*}
\begin{split}
    &P(X_{n+1} = x_{n+1} \mid X_n = x_n, X_{n-1} = x_{n-1}, ..., X_1 = x_1) \\ = &P(X_{n+1} = x_{n+1} \mid X_n = x_n).
\end{split}
\end{align*}
A finite discrete-time Markov chain model is characterized by its transition probability matrix, in which the entries represent the transition probabilities between the set of states.

The structure of the estimated transition matrix $P$ that is applied in oversampling the minority class is illustrated in Table \ref{tab:P}. The matrix is of dimension $(\|V\|+1) \times (\|V\|+1)$, where $\|V\|$ is the size of the entire training vocabulary $V$. We include also a $<$stop$>$ token in the vocabulary to handle word transitions in the beginnings and in the ends of documents. Each row of the matrix corresponds to a distinct word $w_i \in V$, and the entries on the row give the estimated distribution of words following the given word $w_i$.

In Table \ref{tab:P}, the block marked with an asterisk (*) includes word transitions in the minority documents but may also include weighted transitions in the majority documents, as in the majority documents, there can also be subsequent words that both belong to the \emph{minority vocabulary}. Transitions from the \emph{majority-only vocabulary} are allowed only back to the \emph{minority vocabulary} and are sampled from the overall minority class word distribution.

\begin{table}[ht]
\begin{center}
\footnotesize
\begin{tabular}{cccc}
                                                                            & \textbf{\begin{tabular}[c]{@{}c@{}}Minority\\ Vocabulary\end{tabular}} & \textbf{\begin{tabular}[c]{@{}c@{}}Majority-only\\ Vocabulary\end{tabular}} & \textbf{\textless{}stop\textgreater{}}                     \\
\multicolumn{1}{l}{}                                                        & \multicolumn{1}{l}{}                                                   & \multicolumn{1}{l}{}                                                        & \multicolumn{1}{l}{}                                       \\
\textbf{\begin{tabular}[c]{@{}c@{}}Minority\\ Vocabulary\end{tabular}}      & \begin{tabular}[c]{@{}c@{}}($\gamma$-weighted*)\\ word transitions\end{tabular}           & \begin{tabular}[c]{@{}c@{}}$\gamma$-weighted\\ word transitions\end{tabular}       & \begin{tabular}[c]{@{}c@{}}document\\ endings\end{tabular} \\
\multicolumn{1}{l}{}                                                        & \multicolumn{1}{l}{}                                                   & \multicolumn{1}{l}{}                                                        & \multicolumn{1}{l}{}                                       \\
\textbf{\begin{tabular}[c]{@{}c@{}}Majority-only\\ Vocabulary\end{tabular}} & \begin{tabular}[c]{@{}c@{}}minority\\ word distribution\end{tabular}  & 0                                                                           & 0                                                          \\
\multicolumn{1}{l}{}                                                        & \multicolumn{1}{l}{}                                                   & \multicolumn{1}{l}{}                                                        & \multicolumn{1}{l}{}                                       \\
\textbf{\textless{}stop\textgreater{}}                                      & \begin{tabular}[c]{@{}c@{}}initial word\\ distribution\end{tabular}    & 0                                                                           & 0                                                         
\end{tabular}
\caption{The structure of the word transition matrix $P$ that EMCO uses for generating synthetic documents. Bolded texts represent the indices and non-bolded texts correspond to the blocks of the matrix. The element on the $i$th row and $j$th column of $P$ gives the estimated transition frequency $p(w_j \mid w_i)$ for word $w_j$ following after word $w_i$.}\label{tab:P}
\end{center}
\end{table}

Next, we describe the approach for estimating the transition frequency matrix $P$. The training data consists of a set of $n$ text documents $D = (d_1,...,d_n)$ of which documents $D_{\mathrm{min}} = (d_1,...,d_m)$ correspond to the minority class and $D_{\mathrm{maj}} = (d_{m+1},...,d_n)$ to the majority class. The documents are ordered sequences of words in set $V$, where $V$ is the vocabulary of words that appear in the training set. The vocabulary can be expresses as $V = V_{\mathrm{min}} \ \bigcup \ V_{\mathrm{maj-only}}$, where $V_{\mathrm{min}}$ is the \emph{minority vocabulary} and $V_{\mathrm{maj-only}}$ is the \emph{majority-only vocabulary}, and by definition, $V_{\mathrm{min}} \ \bigcap \ V_{\mathrm{maj-only}} = \emptyset$.

For computational convenience, the matrix $P$ is here defined as an unnormalized transition weight matrix, that is, each entry $P(i,j)$ is the estimated transition frequency from word $w_i$ to word $w_j$. Thus, in order to compute the transition probability estimates, each entry should be normalized by dividing it by the sum of the corresponding row.

First, we assign the transitions from the \emph{minority vocabulary} to every word in the total vocabulary, that is, from words $w_i \in V_{\mathrm{min}}$ to words $w_j \in V$, such that
\begin{align*}
    P(i,j) = 
    \begin{cases}
        C_{i,j}^{\mathrm{min}} + \gamma C_{i,j}^{\mathrm{maj}}, \ i \neq j, \\
        0, \ i = j,
    \end{cases}
\end{align*}
where $C_{i,j}^{\mathrm{min}}$ is the count of occurrences of word $w_j$ following word $w_i$ in the minority documents $D_{\mathrm{min}}$, and $C_{i,j}^{\mathrm{maj}}$ is the count of word $w_j$ following word $w_i$ in the majority documents $D_{\mathrm{maj}}$. Note that the transition probabilities from any word back to itself are set to zeros. Hyperparameter $\gamma \geq 0$ is a weight parameter for the transitions in the majority documents. We further discuss the choice of $\gamma$ in Sections \ref{sec:experiments} and \ref{sec:results}.

Next, the transitions from and to $<$stop$>$ token, i.e., the corresponding elements on the last row and the last column of the matrix $P$ in Table \ref{tab:P}, are set such that for words $w_i \in V_{\mathrm{min}}$:
\begin{align*}
    P(l,i) &= I_i^{\mathrm{min}}, \\
    P(i,l) &= E_i^{\mathrm{min}},
\end{align*}
where $I_i^{\mathrm{min}}$ is the number of occurrences of word $w_i$ as the initial word in the minority documents $D_{\mathrm{min}}$, and $E_i^{\mathrm{min}}$ is the count of occurrences of word $w_i$ as the ending word in the minority documents $D_{\mathrm{min}}$, and where $l=\|V\|+1$ is the index of the last row/column of the square matrix $P$.

Finally, the transitions from the \emph{majority-only vocabulary} to the \emph{minority vocabulary} are assigned based on the marginal word distribution of the minority class, that is,
\begin{align*}
    P(i,j) = C_j^{\mathrm{min}},
\end{align*}
where $w_i \in V_{\mathrm{maj-only}}$ and $w_j \in V_{\mathrm{min}}$. The count $C_j^{\mathrm{min}}$ represents the occurrences of word $w_j$ in the minority documents in total. The remaining entries in the matrix $P$ are set to zeros as shown in Table \ref{tab:P}.

A pseudo-code for generating a synthetic minority document following the EMCO approach is shown in Algorithm \ref{Alg1}, where $P$ is the estimated (normalized or unnormalized) transition frequency matrix including the corresponding row/column vocabulary, and $L$ is the minority document length distribution. First, the length of the synthetic document is drawn from $L$. The sequence begins with a $<$stop$>$ token, and each subsequent word is drawn from the distribution given by the row of $P$ corresponding to the current word. Note that $<$stop$>$ tokens are not included in the final document, as their objective is to merely deal with document-ending words. After the transition matrix $P$ has been estimated, the described sampling approach can be applied as many times as needed.

\begin{algorithm}
\caption{Extrapolated Markov Chain Oversampling}\label{Alg1}
\begin{algorithmic}[1]
\Function{Synthetic Document}{P, L}
\State $\textit{length} \gets \text{draw from }\text{L}$
\State $\textit{document} \gets \text{empty vector with a size of }\textit{length}$
\State $\textit{current} \gets \text{$<$stop$>$}$
\State $i \gets 0$
\While {$i < \textit{length}$}
\State $\textit{next} \gets \text{draw from the row of P corresponding to \textit{current}}$
\If {$\textit{next} \neq \text{$<$stop$>$}$}
\State $\textit{document[i]} \gets \textit{next}$
\State $i \gets i+1$
\EndIf
\State $\textit{current} \gets \textit{next}$
\EndWhile
\State \Return $\textit{document}$
\EndFunction
\end{algorithmic}
\end{algorithm}

Note that the Markov chain applied in EMCO is not ergodic. The method utilizes the structure of the language used in both minority and majority class documents, and transition probabilities are based on this empirical structure that depends on the given text type, context, and language. Neither are there any guarantees that the synthetic document generation process by EMCO would provide meaningful sentences. However, the purpose of EMCO is not to generate text, per se, but rather, just as any other oversampling method, to improve generalization when fitting the classifier.

\section{Experiments}
\label{sec:experiments}

We use three different multiclass text data sets for testing the considered oversampling methods: (1) the well-known industry-standard Reuters-21578 corpus (Distribution 1.0, ApteMod version) downloaded from the NLTK Python library \citep{nltk}; (2) HuffPost News Category Dataset \citep{misra2022news}; and (3) the 20 newsgroups data set retrieved via the scikit-learn Python library \citep{scikit-learn}.

Reuters-21578 is a multi-label data set consisting of 21,578 news articles that featured on the Reuters newswire in 1987. Each document has zero, one, or multiple labels. We apply the standard ``ModApte'' split to create training and test sets which ensures that each class has at least one observation both in the training and in the test set. This split accounts for 7,769 training documents and 3,019 test documents. On top of the traditional way of using Reuters-21578, we also generate another set of classification tasks by using just the headlines of the articles.

HuffPost News Category Dataset includes about 210,000 news headlines and abstracts from 2012 to 2022 collected from The Huffington Post's website. The news items are assigned to 42 different categories such that each observation has exactly one label. We create two sets of data from HuffPost News Category Dataset by using (1) just the headlines, and (2) headlines combined with the abstracts.

The 20 newsgroups data set consists of about 18,000 newsgroups posts (i.e., discussion messages) posted on 20 different topics. Each post is assigned exaclty one label, that is, its topic. The set is divided into training and test set based on a specific date, and, after removing data rows with empty documents, there are 10,999 observations in the training set and 7,306 observations in the test set. We also filter topic-related metadata (i.e., headers, signature blocks, and quotation blocks) from the observations in order to obtain more meaningful results.

In each set, the documents are preprocessed by first tokenizing them into tokens that consist of only English letters and then stemming the tokens with the Snowball stemmer \citep{porter80}.\footnote{Note that applying EMCO does not require that the documents are preprocessed, nor does it assume anything about how the documents are modeled in classification as the synthetic observations are generated prior to any data embeddings. However, our experiments show that the benefits of EMCO seem to be most notable with document-embedded approaches.} We remove stopwords from the training corpora by applying an English stopword list included in the NLTK library \citep{nltk}. Moreover, we remove tokens that, after stemming, consist of single characters and tokens that appear only once or twice in the training documents. Empty data rows are dropped after preprocessing.

The average document lengths of each preprocessed training set and the mean shares of minority vocabularies out of total training feature spaces are presented in Table \ref{tab:doc_lengths}. (Note that, as explained in Section \ref{sec:results}, only those categories that would actually get oversampled are selected as minority classes.) The mean document lengths vary from about six tokens up to over a hundred tokens per document and the mean minority vocabulary shares from 4.7\% up to 26.3\%. The variability in data sets helps in evaluating the strengths of the considered oversampling methods on different types of text samples.

\begin{table}[ht]
\footnotesize
\begin{center}
\begin{tabular}{llll}
\textbf{Data set}     & \textbf{Type}   & \textbf{Mean length}	& \textbf{Mean share of minority vocabulary}	\\
\textbf{Reuters}      & Title 		& 5.9			& 4.7\%                       			\\
                      & Full      	& 80.3                  & 10.0\%        				\\
\textbf{HuffPost}     & Title 		& 6.5                   & 20.6\%       					\\
                      & Full      	& 18.3                  & 23.8\%        				\\
\textbf{20Newsgroups} &                 & 108.7                 & 26.3\%
\end{tabular}
\caption{Mean lengths and shares of minority vocabularies of the preprocessed documents.}
\label{tab:doc_lengths}
\end{center}
\end{table}

The documents are modeled as bag-of-words vectors. We transform the vectors with tf-idf weighting \citep[see][]{salton88}, and normalization with tools provided in the scikit-learn library \citep{scikit-learn}. This accounts for two data transformations. First, the term-frequencies of each token $t$ in plain bag-of-words vectors are weighted with a smoothed idf-weight $\text{idf}(t) = \ln \frac{n+1}{\text{df}(t)+1} + 1$, where $n$ is the total number of training documents and $\text{df}(t)$ is the number of documents that include term $t$. After this, the transformed vectors are $L^2$-normalized such that the sum of squares of each vector's elements is equal to one.

We test our method against six other approaches presented in the literature: random oversampling (ROS), SMOTE by \cite{chawla02}, ADASYN by \cite{he08}, Distributional Random Oversampling (DRO) by \cite{moreo16}, DECOM (data re-sampling with
probabilistic topic models) by \cite{chen11}, and oversampling using Easy Data Augmentation (EDA) by \cite{wei19}. As a benchmark, we also train a classifier for each classification task with no oversampling. In addition, as our idea builds on and generalizes a naive Markov chain approach, we include a plain Markov chain oversampling (MCO) approach (i.e., EMCO with $\gamma=0$) in our experiments as well.

For the general-purpose oversampling methods (ROS, SMOTE, and ADASYN) we use implementations in the imbalanced-learn Python toolbox \citep{lemaitre17}. For DRO, we use a Python implementation provided by the authors.\footnote{An implementation of DRO is available at \url{https://github.com/AlexMoreo/pydro}.} For EDA, we also use the original Python implementation by the authors.\footnote{An implementation of EDA is available at \url{https://github.com/jasonwei20/eda_nlp}.} Note that EDA is originally designed for data augmentation, but it can easily be applied for oversampling as well by simply augmenting only minority observations.

We implement DECOM based on \cite{chen11} with scikit-learn library's Latent Dirichlet Allocation (LDA) module which uses batch variational Bayes method for estimation. Following \cite{chen11}, in our implementation of DECOM, we use a uniform Dirichlet prior over the entire vocabulary for each topic's word distribution. Note that, similarly to EMCO, this allows the minority feature space to expand in oversampling, albeit that the expansion in this case is not based on information in the data but rather just on the applied prior.

We use $k=5$ nearest neighbors for SMOTE and ADASYN, following \cite{chawla02} and \cite{he08}. However, if the minority training sample consists of only five or less observations, we use the maximum number of available neighbors. For EMCO, we use $\gamma=1$ and $\gamma=0.1$ (and for MCO, $\gamma=0$). For DECOM, we follow the hyperparameter selection in \cite{chen11} and set the number of topics as $T=30$, the (maximum) number of iterations as $N=300$, and the prior parameters as $\alpha=\frac{50}{T}$ and $\beta=0.01$. EMCO, MCO, and DECOM samples are transformed into tf-idf weighted and normalized bag-of-words space only after oversampling. For simplicity, we use the same transformation pipeline fitted on the original sample for all oversampled training sets as well.

Support vector machine (SVM) with a linear kernel is adopted as the classification algorithm in our experiments. Linear SVM is a computationally light and widely applied classifier that typically performs well in small text classification tasks. We implement the classifier with scikit-learn library's LinearSVC module, which is based on the highly efficient liblinear implementation \citep{liblinear}. We apply standard hyperparameters for SVM, that is, $L^2$-penalized hinge loss, regularization parameter $C=1$, and a stopping tolerance of $10^{-3}$. In addition to SVM, we also experiment our method with a word-embedded neural network classifier.

\section{Results and discussion}
\label{sec:results}

In this section, we present the results of empirical testing of the considered oversampling methods.\footnote{An implementation of the introduced method, source code of the experiments, and information about accessing the data sets are available at \url{https://github.com/AleksiAvela/emco}.} As oversampling and fitting the classifier include some randomness, oversampling and classification is repeated multiple times for each task and the category level results are averages of these repetitions. For the Reuters-21578 articles and headlines and for the 20 newsgroups posts classification is repeated five times for each category. We use the built-in train-test splits in these data sets. For the HuffPost headlines (where there are almost 200,000 documents) classification is done only once for each category with a random train-test split such that half of the data is used for training and the remaining half for testing. In addition, we randomly split the data rows of HuffPost data set into five same-sized subsets (and randomly divide them half-and-half into training and test sets) and evaluate classification on these sets separately for the headlines and abstracts.

Although imbalanced classes are often oversampled to full balance, studies have shown that smaller oversampling ratios may sometimes be preferable (see, e.g., \cite{piyadasa23}). Moreover, the optimal oversampling ratio depends on the domain, the applied classification algorithms, and the objectives of the task \citep{piyadasa23}. For instance, \cite{moreo16} use sampling ratios ranging from 5\% to 20\%. Based on this, we conduct our experiments by applying two oversampling ratios, 10\% and 20\%, to each classification task. That is, after oversampling, the relative frequency of minority class in each training set is 10\% or 20\%, respectively.

We select only those categories to be considered as minority classes that, given the oversampling ratio, would actually get oversampled. We set the threshold such that a category is considered as a minority class if its relative frequency in the training set is less than 0.75 $\times$ sampling ratio. We were not able to apply DRO for the full HuffPost data set due to a memory error and thus we do not report any results for DRO in that particular experiment.

Our main interest is on three evaluation statistics: balanced accuracy (BA), and $F_\beta$-score with $\beta=1$ and $\beta=2$. Balanced accuracy is the average of recall (i.e., true positive rate or tpr) and true negative rate (tnr). $F_1$-score is the harmonic mean of recall and precision (the number of true positives divided by the total number of positive predictions). $F_2$-score is similar to $F_1$-score, but it gives twice as much weight to recall as to precision. $F_2$-score is commonly applied in evaluation of imbalanced classification as it acknowledges the fact that, when dealing with imbalanced data, usually the cost of a false negative is much higher than the cost of a false positive. Equations of all evaluation measures applied in this work are listed in Appendix \ref{appendix:measures}.

Macro-averaged balanced accuracies and $F_1$- and $F_2$-scores obtained in the experiments are presented in Tables \ref{tab:bacc}, \ref{tab:F1}, and \ref{tab:F2}, respectively. Note that if there are no positive (negative) observations in the test set, then recall (tnr) is not defined. This is not an issue, though, as in all categories of the considered test sets, there is at least one observation in both classes. However, precision is not defined if the classifier does not produce any positive predictions, which can often happen with imbalanced data. In these cases, we set the value of precision to zero. Obtained recall, tnr, and precision values are presented in Appendix \ref{appendix:tables}.

We have aggregated the results separately for low frequency categories (whose relative frequency in the training set is higher than or equal to 1.5\%) and for very low frequency categories (relative frequency lower than 1.5\%). However, note that in 20Newsgroups data set, every category has a relative frequency higher than 1.5\% in the training set, and thus there are no results for very low frequency categories for 20Newsgroups. The columns of the tables account for the different approaches, where ``SVM'' is the benchmark linear SVM classifier trained on the original data with no oversampling. The top three performers on each row are marked with respective numbers and the best result is also bolded.

In balanced accuracy, EMCO with $\gamma=1$ outperforms the other approaches in very low frequency categories on every row, except for one, where EMCO with $\gamma=0.1$ is the top performer. In low frequency categories, there are two cases in which EMCO is not in the top three with neither $\gamma=1$ nor $\gamma=0.1$. However, in general, EMCO produces excellent results in low frequency categories as well, particularly with the higher value of $\gamma$. The overall great performance of EMCO in balanced accuracy is explained by its ability to generate a high recall without compromising tnr too much, as shown in Appendix \ref{appendix:tables}. Moreover, in particular in very low frequency categories, the higher sampling ratio (i.e., 20\%) produces better performance compared to the lower sampling ratio (i.e., 10\%). Naturally, it seems that higher oversampling ratio improves recall.

\begin{table}[ht]
\begin{center}
\scriptsize
\begin{tabular}{llllllllllll}
\\ \multicolumn{4}{l}{\scriptsize{\textbf{Very low frequency categories}}} & & & & & & & & \\
\multicolumn{2}{l}{\textbf{Sampling ratio: 10\% }} & SVM & ADASYN & DECOM & DRO & EDA & EMCO & EMCO & MCO & ROS & SMOTE \\
 & & & & & & & $\gamma$=1 & $\gamma$=0.1 & & & \\ \hline
\textbf{Reuters$^\dagger$} & \textbf{Titles} & .619 & .684 & .704$^3$ & .687 & .682 & {\bf .736}$^1$ & .734$^2$ & .679 & .683 & .684 \\
                             & \textbf{Full} & .656 & .713 & .768$^3$ & .731 & .720 & {\bf .801}$^1$ & .791$^2$ & .721 & .713 & .713 \\
\textbf{HuffPost}          & \textbf{Titles*} & .515 & .606 & .613 & .606 & .615$^3$ & {\bf .638}$^1$ & .630$^2$ & .601 & .600 & .605 \\
                             & \textbf{Titles} & .533 & .701$^3$ & .693 & NA & .706$^2$ & .701 & {\bf .709}$^1$ & .686 & .689 & .699 \\
                             & \textbf{Full*} & .517 & .591 & .618$^3$ & .598 & .603 & {\bf .663}$^1$ & .658$^2$ & .589 & .591 & .591 \\
\multicolumn{2}{l}{\textbf{Sampling ratio: 20\% }} & & & & & & & & & & \\ \hline
\textbf{Reuters$^\dagger$} & \textbf{Titles} & .619 & .683 & .716$^3$ & .688 & .678 & {\bf .750}$^1$ & .746$^2$ & .681 & .683 & .683 \\
                             & \textbf{Full} & .656 & .713 & .775$^3$ & .733 & .717 & {\bf .815}$^1$ & .809$^2$ & .730 & .713 & .713 \\
\textbf{HuffPost}          & \textbf{Titles*} & .515 & .612 & .661$^3$ & .609 & .626 & {\bf .683}$^1$ & .667$^2$ & .616 & .602 & .612 \\
                             & \textbf{Titles} & .533 & .712 & .736$^3$ & NA & .725 & {\bf .772}$^1$ & .771$^2$ & .712 & .695 & .712 \\
                             & \textbf{Full*} & .517 & .591 & .652$^3$ & .601 & .613 & {\bf .722}$^1$ & .709$^2$ & .606 & .591 & .591 \\
\multicolumn{4}{l}{\scriptsize{\textbf{Low frequency categories}}} & & & & & & & & \\
\multicolumn{2}{l}{\textbf{Sampling ratio: 10\% }} & & & & & & & & & & \\ \hline
\textbf{Reuters$^\dagger$} & \textbf{Titles} & .790 & .841 & .845 & .844 & .848$^2$ & {\bf .851}$^1$ & .846$^3$ & .836 & .838 & .837 \\
                             & \textbf{Full} & .858 & .897 & .901$^3$ & .898 & .900 & .916$^2$ & {\bf .919}$^1$ & .898 & .890 & .892 \\
\textbf{HuffPost}          & \textbf{Titles*} & .611 & .709$^3$ & .695 & .711$^2$ & {\bf .721}$^1$ & .695 & .700 & .693 & .706 & .708 \\
                             & \textbf{Titles} & .651 & .765$^3$ & .759 & NA & {\bf .781}$^1$ & .726 & .745 & .758 & .771$^2$ & .763 \\
                             & \textbf{Full*} & .622 & .714 & .711 & .717$^3$ & {\bf .726}$^1$ & .706 & .720$^2$ & .695 & .711 & .710 \\
\multicolumn{2}{r}{\textbf{20Newsgroups$^\dagger$}} & .737 & .774 & .777$^3$ & {\bf .779}$^1$ & .756 & .770 & .778$^2$ & .771 & .771 & .772 \\
\multicolumn{2}{l}{\textbf{Sampling ratio: 20\% }} & & & & & & & & & & \\ \hline
\textbf{Reuters$^\dagger$} & \textbf{Titles} & .790 & .852 & .878$^2$ & .856 & .864 & {\bf .883}$^1$ & .877$^3$ & .862 & .851 & .848 \\
                             & \textbf{Full} & .858 & .899 & .912$^3$ & .907 & .906 & {\bf .950}$^1$ & .946$^2$ & .909 & .896 & .898 \\
\textbf{HuffPost}          & \textbf{Titles*} & .614 & .737 & .746$^3$ & .733 & .743 & {\bf .751}$^1$ & .750$^2$ & .725 & .727 & .736 \\
                             & \textbf{Titles} & .656 & .799 & .809$^2$ & NA & {\bf .814}$^1$ & .790 & .806$^3$ & .799 & .802 & .797 \\
                             & \textbf{Full*} & .628 & .729 & .757$^3$ & .739 & .747 & .779$^2$ & {\bf .782}$^1$ & .729 & .728 & .728 \\
\multicolumn{2}{r}{\textbf{20Newsgroups$^\dagger$}} & .737 & .794 & .809$^3$ & .799 & .769 & .818$^2$ & {\bf .823}$^1$ & .804 & .788 & .791 \\
\end{tabular}
\\
\scriptsize{$^\dagger$Averages of 5 repetitions per category. $^*$Averages of 5-fold split of the data set.}
\caption{Average balanced accuracy.}
\label{tab:bacc}
\end{center}
\end{table}

All the methods designed for text oversampling (i.e., DECOM, DRO, EDA, and EMCO) perform better with respect to balanced accuracy than the rest of the approaches. In low frequency categories of larger data sets, EDA seems to perform particularly well. However, even the general-purpose oversampling approaches seem to improve the performance of the classifier compared to no oversampling. In addition, it seems that EMCO performs exceptionally well in cases where there are only very limited amounts of training data. This is likely caused by the fact that EMCO's ability to incorporate information from the majority class is particularly useful when there is only very little information available in the minority class. On the other hand, it may be that in the larger data sets, EMCO tries to force too much information from the majority documents, and therefore its advantage compared to other approaches is diminished.

DRO and, in particular, EDA, seem to be the overall top performers with respect to $F_1$-score in very low frequency categories. Notably, MCO (i.e., EMCO with $\gamma=0$) and EMCO with $\gamma=0.1$ perform better with respect to $F_1$-score than EMCO with $\gamma=1$ and are the top performers in multiple cases. This is probably due to the fact that, with a high value of $\gamma$, EMCO forces a high recall on the expense of precision. This highlights the trade-off between recall and precision with EMCO, which can be controlled with the hyperparameter $\gamma$.

\begin{figure}[ht]
\centering
\includegraphics[width=0.82\textwidth]{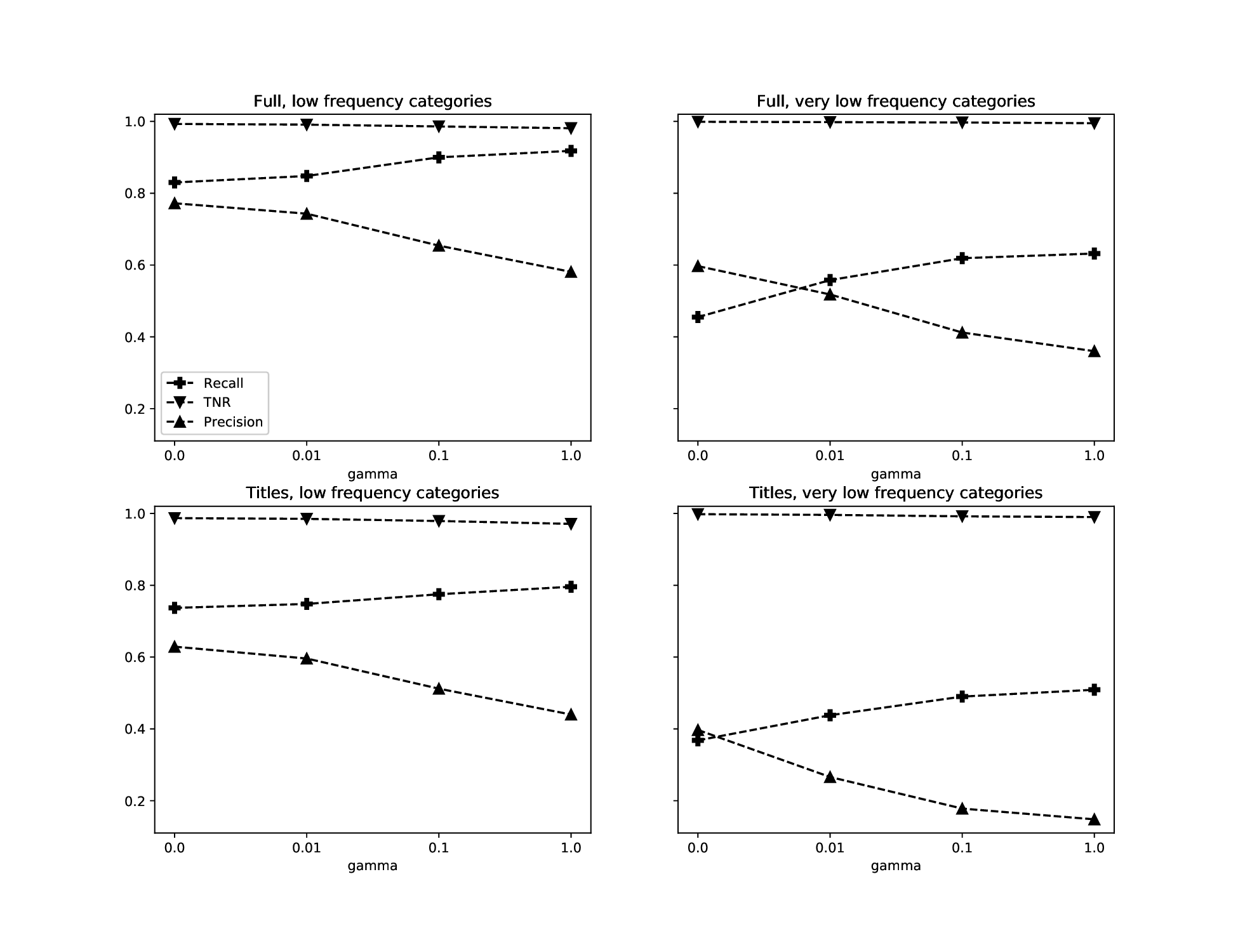}
\caption{The effect of EMCO's $\gamma$-hyperparameter on recall, tnr, and precision with Reuters data set based on averages of five repetitions per category with a sampling ratio of 20\%. Note that the values on the horizontal axis are not distributed linearly.}\label{fig:gamma_reuters}
\end{figure}

The effect of $\gamma$ on recall, tnr, and precision is also illustrated in Figure \ref{fig:gamma_reuters}, where we have reported macro-averaged results on Reuters articles and titles with $\gamma \in (0.0, 0.01, 0.1, 1.0)$ with sampling ratio of 20\%. The findings seem intuitive; the higher the value of $\gamma$, the more words from the \emph{majority-only vocabulary} are included in the synthetic documents, which, in turn, forces the decision boundary closer to the negative class. This causes the number of true positives to increase, but, due to class-overlapping, it also increases the number of false positives.

Even small values of $\gamma$ seem to improve the classification performance with respect to recall and balanced accuracy compared to $\gamma=0$. The optimal choice of $\gamma$ depends on the context and the relative importance between recall and precision and/or tnr in the given application. We would not set $\gamma$ higher than one, but, for instance, values between $(0, 0.1]$ could be experimented with in cases where, e.g., $F_1$- or $F_2$-score is more important performance statistic than balanced accuracy, as lower values of $\gamma$ likely result in a higher precision.

\begin{table}[ht]
\begin{center}
\scriptsize
\begin{tabular}{llllllllllll}
\\ \multicolumn{4}{l}{\scriptsize{\textbf{Very low frequency categories}}} & & & & & & & & \\
\multicolumn{2}{l}{\textbf{Sampling ratio: 10\% }} & SVM & ADASYN & DECOM & DRO & EDA & EMCO & EMCO & MCO & ROS & SMOTE \\
 & & & & & & & $\gamma$=1 & $\gamma$=0.1 & & & \\ \hline
\textbf{Reuters$^\dagger$} & \textbf{Titles} & .300 & .405$^3$ & .226 & {\bf .407}$^1$ & .375 & .277 & .308 & .374 & .404 & .406$^2$ \\
                             & \textbf{Full} & .376 & .487 & .449 & .512$^2$ & .491 & .502$^3$ & {\bf .515}$^1$ & .495 & .487 & .487 \\
\textbf{HuffPost}          & \textbf{Titles*} & .054 & .224$^2$ & .206 & .221 & {\bf .227}$^1$ & .207 & .212 & .217 & .216 & .224$^3$ \\
                             & \textbf{Titles} & .109 & .303$^3$ & .279 & NA & .301 & .278 & .296 & {\bf .312}$^1$ & .293 & .305$^2$ \\
                             & \textbf{Full*} & .059 & .240 & .242 & .250 & .256$^3$ & .270$^2$ & {\bf .279}$^1$ & .238 & .240 & .240 \\
\multicolumn{2}{l}{\textbf{Sampling ratio: 20\% }} & & & & & & & & & & \\ \hline
\textbf{Reuters$^\dagger$} & \textbf{Titles} & .300 & .404$^3$ & .123 & {\bf .407}$^1$ & .346 & .219 & .250 & .361 & .403 & .404$^2$ \\
                             & \textbf{Full} & .376 & .487$^3$ & .390 & {\bf .517}$^1$ & .482 & .436 & .465 & .499$^2$ & .487 & .487 \\
\textbf{HuffPost}          & \textbf{Titles*} & .054 & .218$^3$ & .144 & .216 & {\bf .220}$^1$ & .161 & .169 & .203 & .213 & .218$^2$ \\
                             & \textbf{Titles} & .109 & .267 & .203 & NA & {\bf .270}$^1$ & .207 & .222 & .269$^3$ & .269$^2$ & .268 \\
                             & \textbf{Full*} & .059 & .240 & .216 & .253$^3$ & {\bf .265}$^1$ & .217 & .235 & .255$^2$ & .240 & .240 \\
\multicolumn{4}{l}{\scriptsize{\textbf{Low frequency categories}}} & & & & & & & & \\
\multicolumn{2}{l}{\textbf{Sampling ratio: 10\% }} & & & & & & & & & & \\ \hline
\textbf{Reuters$^\dagger$} & \textbf{Titles} & .695 & .696 & .670 & .687 & .693 & .653 & .681 & .700$^3$ & .701$^2$ & {\bf .709}$^1$ \\
                             & \textbf{Full} & .792 & .810 & .810 & .810 & {\bf .815}$^1$ & .785 & .802 & .814$^2$ & .809 & .813$^3$ \\
\textbf{HuffPost}          & \textbf{Titles*} & .303 & .455$^2$ & .437 & .429 & {\bf .455}$^1$ & .430 & .444 & .440 & .446 & .455$^3$ \\
                             & \textbf{Titles} & .387 & .517 & .523 & NA & {\bf .532}$^1$ & .478 & .507 & .529$^2$ & .526$^3$ & .520 \\
                             & \textbf{Full*} & .325 & .480$^3$ & .473 & .473 & {\bf .489}$^1$ & .462 & .483$^2$ & .466 & .476 & .479 \\
\multicolumn{2}{r}{\textbf{20Newsgroups$^\dagger$}} & .602 & .654$^3$ & {\bf .658}$^1$ & .650 & .617 & .645 & .657$^2$ & .650 & .649 & .650 \\
\multicolumn{2}{l}{\textbf{Sampling ratio: 20\% }} & & & & & & & & & & \\ \hline
\textbf{Reuters$^\dagger$} & \textbf{Titles} & {\bf .695}$^1$ & .673 & .588 & .667 & .664 & .551 & .611 & .673$^3$ & .673 & .693$^2$ \\
                             & \textbf{Full} & .792 & .795 & .770 & .804$^2$ & .801 & .700 & .754 & .792 & .801$^3$ & {\bf .815}$^1$ \\
\textbf{HuffPost}          & \textbf{Titles*} & .314 & .447$^2$ & .420 & .432 & .442$^3$ & .411 & .428 & .441 & .437 & {\bf .447}$^1$ \\
                             & \textbf{Titles} & .402 & .475 & .492 & NA & .503$^2$ & .462 & .496 & {\bf .514}$^1$ & .500$^3$ & .480 \\
                             & \textbf{Full*} & .342 & .489 & .483 & .490 & {\bf .498}$^1$ & .467 & .487 & .494$^2$ & .487 & .490$^3$ \\
\multicolumn{2}{r}{\textbf{20Newsgroups$^\dagger$}} & .602 & .664 & .676$^3$ & .662 & .626 & .660 & .677$^2$ & {\bf .679}$^1$ & .656 & .661 \\
\end{tabular}
\\
\scriptsize{$^\dagger$Averages of 5 repetitions per category. $^*$Averages of 5-fold split of the data set.}
\caption{Macro-averaged F1-score.}
\label{tab:F1}
\end{center}
\end{table}

$F_2$-score assigns a greater weight on recall compared to precision than $F_1$-score, which is often reasonable when dealing with imbalanced classes and misclassification costs, as mentioned, for instance, by \cite{köknar09} and \cite{yin20}. Especially in very low frequency categories, EMCO produces excellent results with respect to $F_2$-score and is the top performer on multiple rows with both $\gamma=1$ and $\gamma=0.1$. In low frequency categories, EDA seems to perform well, but with the higher sampling ratio, EMCO with $\gamma=0.1$ produces excellent results as well. The general-purpose oversampling approaches also perform reasonably well when considering the $F_\beta$-scores. This is due to the fact that, while they lose in recall to the considered text oversampling methods, they are able to compensate this with a higher precision. The performance of EMCO is better in $F_2$-score than in $F_1$-score, which is due to its superior ability to obtain a high recall. Namely, with respect to average recall, EMCO outperforms the other considered approaches, especially in very low frequency categories.

\begin{table}[ht]
\begin{center}
\scriptsize
\begin{tabular}{llllllllllll}
\\ \multicolumn{4}{l}{\scriptsize{\textbf{Very low frequency categories}}} & & & & & & & & \\
\multicolumn{2}{l}{\textbf{Sampling ratio: 10\% }} & SVM & ADASYN & DECOM & DRO & EDA & EMCO & EMCO & MCO & ROS & SMOTE \\
 & & & & & & & $\gamma$=1 & $\gamma$=0.1 & & & \\ \hline
\textbf{Reuters$^\dagger$} & \textbf{Titles} & .258 & .380$^3$ & .286 & {\bf .385}$^1$ & .366 & .358 & .376 & .363 & .379 & .380$^2$ \\
                             & \textbf{Full} & .333 & .447 & .478 & .479$^3$ & .458 & {\bf .547}$^1$ & .544$^2$ & .459 & .446 & .446 \\
\textbf{HuffPost}          & \textbf{Titles*} & .037 & .220 & .221 & .219 & .233$^3$ & {\bf .246}$^1$ & .242$^2$ & .211 & .210 & .219 \\
                             & \textbf{Titles} & .078 & .359$^3$ & .338 & NA & {\bf .363}$^1$ & .343 & .361$^2$ & .349 & .343 & .358 \\
                             & \textbf{Full*} & .040 & .202 & .242$^3$ & .217 & .225 & {\bf .304}$^1$ & .303$^2$ & .200 & .202 & .202 \\
\multicolumn{2}{l}{\textbf{Sampling ratio: 20\% }} & & & & & & & & & & \\ \hline
\textbf{Reuters$^\dagger$} & \textbf{Titles} & .258 & .379$^3$ & .200 & {\bf .386}$^1$ & .350 & .322 & .345 & .360 & .379 & .379$^2$ \\
                             & \textbf{Full} & .333 & .446 & .447 & .484$^3$ & .451 & .520$^2$ & {\bf .533}$^1$ & .473 & .446 & .446 \\
\textbf{HuffPost}          & \textbf{Titles*} & .037 & .224 & .219 & .220 & .242$^3$ & {\bf .247}$^1$ & .245$^2$ & .224 & .210 & .224 \\
                             & \textbf{Titles} & .078 & .346 & .311 & NA & {\bf .360}$^1$ & .331 & .345 & .348$^2$ & .335 & .347$^3$ \\
                             & \textbf{Full*} & .040 & .202 & .266$^3$ & .221 & .241 & .317$^2$ & {\bf .323}$^1$ & .229 & .202 & .202 \\
\multicolumn{4}{l}{\scriptsize{\textbf{Low frequency categories}}} & & & & & & & & \\
\multicolumn{2}{l}{\textbf{Sampling ratio: 10\% }} & & & & & & & & & & \\ \hline
\textbf{Reuters$^\dagger$} & \textbf{Titles} & .622 & .692$^2$ & .685 & .692$^3$ & {\bf .700}$^1$ & .686 & .692 & .687 & .690 & .691 \\
                             & \textbf{Full} & .747 & .804 & .808 & .805 & .809$^3$ & .814$^2$ & {\bf .826}$^1$ & .805 & .794 & .798 \\
\textbf{HuffPost}          & \textbf{Titles*} & .250 & .439$^2$ & .413 & .433 & {\bf .456}$^1$ & .410 & .423 & .411 & .433 & .437$^3$ \\
                             & \textbf{Titles} & .332 & .531$^3$ & .525 & NA & {\bf .557}$^1$ & .467 & .502 & .528 & .543$^2$ & .530 \\
                             & \textbf{Full*} & .271 & .454 & .446 & .456$^3$ & {\bf .473}$^1$ & .435 & .461$^2$ & .422 & .448 & .447 \\
\multicolumn{2}{r}{\textbf{20Newsgroups$^\dagger$}} & .519 & .591 & .595$^3$ & {\bf .597}$^1$ & .554 & .581 & .597$^2$ & .584 & .584 & .586 \\
\multicolumn{2}{l}{\textbf{Sampling ratio: 20\% }} & & & & & & & & & & \\ \hline
\textbf{Reuters$^\dagger$} & \textbf{Titles} & .622 & .697 & .683 & .700 & .708$^2$ & .668 & .695 & {\bf .708}$^1$ & .696 & .700$^3$ \\
                             & \textbf{Full} & .747 & .800 & .805 & .813$^2$ & .812$^3$ & .809 & {\bf .834}$^1$ & .810 & .799 & .806 \\
\textbf{HuffPost}          & \textbf{Titles*} & .257 & .475$^3$ & .472 & .465 & .480$^2$ & .472 & {\bf .480}$^1$ & .458 & .459 & .473 \\
                             & \textbf{Titles} & .344 & .553 & .571$^3$ & NA & {\bf .582}$^1$ & .535 & .569 & .572$^2$ & .567 & .554 \\
                             & \textbf{Full*} & .286 & .479 & .512$^3$ & .493 & .506 & .527$^2$ & {\bf .541}$^1$ & .480 & .477 & .478 \\
\multicolumn{2}{r}{\textbf{20Newsgroups$^\dagger$}} & .519 & .623 & .648$^3$ & .629 & .577 & .655$^2$ & {\bf .666}$^1$ & .641 & .611 & .618 \\
\end{tabular}
\\
\scriptsize{$^\dagger$Averages of 5 repetitions per category. $^*$Averages of 5-fold split of the data set.}
\caption{Macro-averaged F2-score.}
\label{tab:F2}
\end{center}
\end{table}

As shown in Table \ref{tab:F1}, when the classification performance is evaluated using $F_1$-score, it seems that in a few cases, classification without sampling performs surprisingly well when compared to classification with oversampling. However, when the evaluation is based on balanced accuracy, classification with oversampling leads to better performance, as shown in Table \ref{tab:bacc}. In particular, EMCO outperforms most of its competitors. This is noteworthy, as balanced accuracy is often regarded as a particularly suitable measure for tasks with highly imbalanced classes and costs (see, e.g., \cite{avela24,henning23}).

With respect to most of the considered evaluation statistics, EMCO seems to produce better results when the imbalance is more severe and/or the training set in general is small. This highlights the importance of EMCO's main feature of including information also outside of the minority class in the oversampled set in cases where information in the minority training set is inherently limited. In our experiments, all the oversampling methods are, for the most part, able to outperform the baseline SVM trained on the original data (except for true negative rate and precision), highlighting the well-known positive effect of oversampling in imbalanced classification.

\begin{table}[ht]
\begin{center}
\scriptsize
\begin{tabular}{lllllllll}
\multicolumn{3}{c}{\textbf{Very low frequency categories}}                & \textbf{BA} & \textbf{F1} & \textbf{F2} & \textbf{TPR} & \textbf{TNR} & \textbf{Precision} \\
                         &                         & LSTM                 & .603              & .246       & .219       & .207        & .999        & .394              \\
\hline
\multicolumn{2}{c}{Sampling ratio = 10\%} & EDA                  & .680              & .289       & .320       & .363        & .996        & .277              \\
                         &                         & EMCO                 & .687              & .179       & .250       & .384        & .991        & .128              \\
\hline
\multicolumn{2}{c}{Sampling ratio = 20\%} & EDA                  & .692              & .294       & .333       & .390        & .995        & .268              \\
                         &                         & EMCO                 & .705              & .161       & .243       & .424        & .986        & .105              \\
\multicolumn{3}{c}{\textbf{Low frequency categories}}                     &                    &             &             &              &              &                    \\
                         &                         & LSTM                 & .797              & .661       & .624       & .602        & .993        & .745              \\
\hline
\multicolumn{2}{c}{Sampling ratio = 10\%} & EDA                  & .839              & .582       & .642       & .699        & .980        & .515              \\
                         &                         & EMCO                 & .840              & .602       & .646       & .697        & .982        & .579              \\
\hline
\multicolumn{2}{c}{Sampling ratio = 20\%} & EDA                  & .870              & .573       & .670       & .770        & .969        & .471              \\
                         &                         & EMCO                 & .852              & .542       & .635       & .735        & .968        & .447             
\end{tabular}
\caption{Macro-averaged results for Reuters titles with an LSTM neural network.}
\label{tab:lstm}
\end{center}
\end{table}

We also run experiments on an LSTM neural network \citep{hochreiter97} with fastText word embeddings\footnote{Retrieved on 13.12.2024 from \url{https://fasttext.cc/docs/en/english-vectors.html}} pre-trained on Wikipedia 2017, UMBC webbase corpus and statmt.org news dataset \citep{mikolov18}. The network consists of a bidirectional LSTM layer (with an output dimension of 32), a fully connected layer with ReLU activation (output dimension of 32), and a 0.4 dropout rate before a final sigmoid activated output node. We train the network for each class of the Reuters headlines data set with three epochs and binary cross-entropy loss. The applied fastText vocabulary includes one million word embeddings with an embedding length of 300.

The macro-averaged classification results (separately for low and very low frequency categories) with the neural network alone as well as combined with EMCO and EDA oversampling are shown in Table \ref{tab:lstm}. It is clear that, in this simple example, EMCO is able to improve classification performance also with a neural network classifier. Again, EMCO proves to be particularly useful when the minority classes have very low frequencies and when the cost of false negatives far surpasses the cost of false positives (i.e., when tpr and balanced accuracy (BA) are of high importance).

The issue of EMCO with sequence-based classification algorithms can be that, as EMCO does not consider long-term dependencies between words in the training documents, longer EMCO documents may not produce the best possible generalization. Thus, it seems that EMCO provides the highest advantage when texts are modeled with a document-embedded approach. Yet, in many statistics reported for the word-embedded example, the performance of EMCO is at least on a similar level when compared to EDA.

The main objective of EMCO is to tackle the issue of convex minority feature space in oversampling without relying on any external sources (such as pre-trained language models or synonym dictionaries). That is, EMCO is independent of the language that it is applied to, and also, independent of any prior work on, e.g., pre-trained word embeddings. The strength of our approach lies in the fact that it can be applied in almost every text classification case; this is particularly important if we are working with limited training data sets (or, e.g., with rare languages).

\section{Oversampling and synthetic vocabulary growth}
\label{sec:vocabulary}

One key aspect in common between the text oversampling methods considered here is that, instead of just ``re-weighting'' the minority class, they also seek to add more useful information to the synthetic sample. The aim of DRO, on top of oversampling, is to assist the classifier generalize the minority class by considering the unique nature of text data via the transformation to the latent space. DECOM approaches oversampling with a semantic topic-model, where the applied flat topic-word prior, similarly to EMCO, enables the minority feature space to increase in oversampling. On the other hand, with EDA, the minority vocabulary is expanded by the synonym replacement procedure.

The minority feature space expansion can itself be regarded as a binary classification task where the words in the \emph{majority-only vocabulary} are the observations which get a positive label if the given word appears in the minority test observations and otherwise a negative label. Based on this idea, we evaluate the differences between how EMCO, DECOM, and EDA are able to increase the minority feature space. An example of this with Reuters titles and 20\% sampling ratio is shown in Table \ref{tab:syn_vocabulary}, where we have reported macro-averaged recall, true negative rate, balanced accuracy, and the number of new synthetic words for all the categories (with five repetitions per one class).

\begin{table}[ht]
\footnotesize
\begin{center}
\begin{tabular}{lllll}
               & \textbf{Recall} & \textbf{TNR} & \textbf{Balanced accuracy} & \textbf{Synthetic words} \\
\textbf{EMCO ($\gamma=1$)}  & 0.57            & 0.80         & 0.69  & 465.68                     \\
\textbf{EMCO ($\gamma=0.1$)}  & 0.55            & 0.81         & 0.68  & 435.85                     \\
\textbf{EDA}  & 0.09            & 0.97         & 0.53  & 71.55                  \\
\textbf{DECOM} & 0.93            & 0.10         & 0.51   & 2040.21                  
\end{tabular}
\caption{Macro-averaged binary classification statistics of EMCO ($\gamma=1$ and $\gamma=0.1$), EDA, and DECOM generated synthetic vocabularies compared to the true minority feature space expansion in the Reuters titles test set. The last column displays the average number of new words in the oversampled minority vocabulary.}
\label{tab:syn_vocabulary}
\end{center}
\end{table}

When EMCO, EDA, and DECOM are compared, there is a large variation in the number of new words included in the respective oversampled minority vocabularies. With EDA, the vocabulary growth is the smallest and tnr is the highest. Using DECOM leads to a larger synthetic minority vocabulary when compared to using EMCO. This results in a higher recall in the DECOM synthetic vocabulary, but a low tnr. With respect to $\gamma$, EMCO behaves as expected. A larger $\gamma$ leads to a larger number of new words and a slightly higher recall. In terms of balanced accuracy, the synthetic vocabulary generated by EMCO produces better results than EDA and DECOM. This is not surprising, as the feature space growth with EDA is based on synonym replacement, and with DECOM, it is based on the uniform topic-word prior.  EMCO generates synthetic vocabulary based on the sequential information in the data.

\begin{figure}[ht]
\centering
\includegraphics[width=0.66\textwidth]{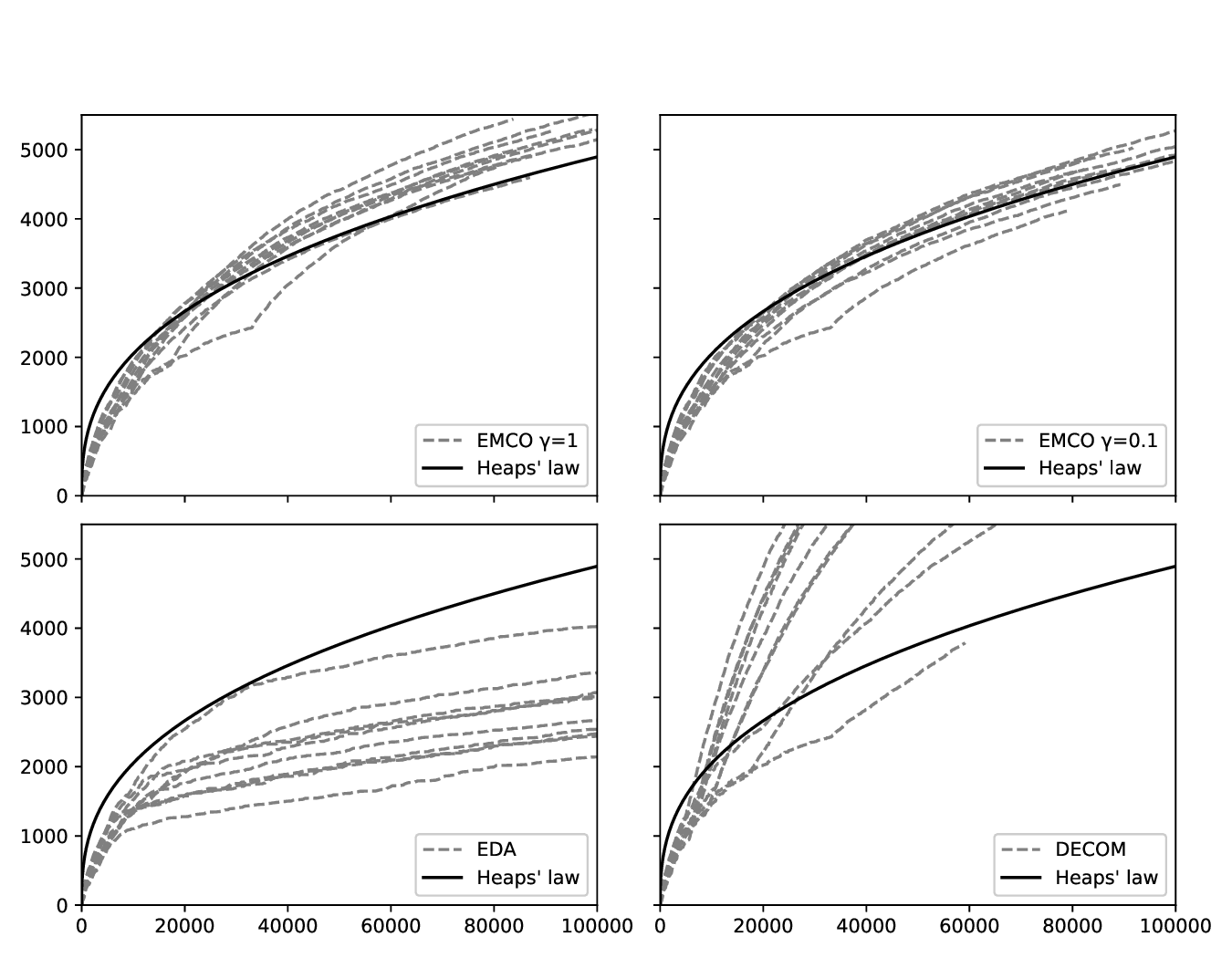}
\caption{Synthetic vocabulary growth with different sampling methods for ten randomly selected Reuters categories.}\label{Fig:heapslaw}
\end{figure}

Another way to consider an oversampling method’s ability to replicate natural vocabulary growth in oversampling is to compare it to Heaps’ law. Application of Heaps’ law to different languages and corpora has been widely studied, and it can generally be regarded as an accurate approximation of how the size of the vocabulary should grow when the sample size increases (see, e.g., \cite{sano12}). As a form of validation for our approach, we compare the minority vocabulary growth predicted by Heaps’ law to what is achieved by applying different oversampling methods. In Figure \ref{Fig:heapslaw}, the size of the (synthetic) minority vocabulary achieved with different oversampling methods is plotted with dashed curves for ten randomly selected Reuters categories against the total number of words in the (synthetic) minority sample. As a reference, the solid curve represents the Heaps’ law fitted on the full training data. The corresponding estimated parameter values are $k \approx 63$ and $\theta \approx 0.378$.

In EDA, the synthetic vocabulary growth is due to the synonym replacement procedure, whereas in DECOM, it is due to the uniform prior for the vocabulary. Note that for any convex oversampling method the dashed curves would simply be constant lines after oversampling. On the other hand, with EMCO, the synthetic vocabulary growth is very close to what is predicted by Heaps’ law. In this example, with $\gamma = 1$, the growth rate slightly overshoots Heaps’ law, and with $\gamma = 0.1$, the growth rate seems to match Heaps’ law rather well. Yet, it seems that EMCO is not too sensitive to the hyperparameter $\gamma$.

\section{Conclusion}
\label{sec:conclusion}

In this work, we discussed the problem of imbalanced data in general as well as in the context of text data. We highlighted issues related to approaching imbalanced text classification with general-purpose oversampling methods. Motivated by this, we introduced a novel approach for text oversampling called Extrapolated Markov Chain Oversampling method---EMCO. Our method is based on modeling the minority documents as a Markov chain, in which the transitions probabilities are estimated by incorporating information also form the majority class. The main idea is to take into account one of the most notable distinctive properties of text data, namely, that the feature space in text data often does not remain immutable when the sample size increases.

We presented our approach based on a framework of modeling text as a combination of two structures: topic and sequence, which are assumed to be partly independent. Based on this idea, we developed the novel oversampling method such that it allows the synthetic minority feature space to expand in oversampling. We discussed some prominent oversampling methods presented in the prior literature, designed for both general purposes as well as for text data. Finally, we evaluated the considered methods on multiple well-known and widely-applied multiclass text data sets with a linear SVM as the learning algorithm. In addition, we tested our oversampling method with a word-embedded neural network classifier as well.

The results of this work are in line with the prior literature. That is, (1) in general, oversampling improves classification performance on imbalanced data, and (2) oversampling methods designed for text data produce overall better performance than general-purpose approaches in text classification tasks. Moreover, we showed that in certain statistics, particularly in recall and balanced accuracy, our novel method is able to outperform the considered counterparts, especially when the class imbalance is severe. We also demonstrated the flexibility of our method with hyperparameter selection. The purpose of the one hyperparameter of EMCO is very intuitive, and we also illustrated how it can be used for controlling the trade-off between recall and tnr/precision.

It is clear that, as EMCO is based on fitting a transition probability matrix of a size of the vocabulary length squared, there can be computational and memory scalability issues. However, there are computational tricks that help with these issues; in particular, the transition matrix is sparse, and thus in our implementation, we only store the indices and weights that correspond to non-zero entries. On top of that, as the experiments show, the strength of our approach lies especially in tasks with limited training data. A possible future prospect for our method would be to experiment with an $n$-gram model instead of the applied 1-gram Markov chain model in order to allow sampling to consider more long-range dependencies between words in a text.

As an oversampling method, EMCO offers more than just a re-weighting of the classes, as it is able to access information in the data that could not be considered with general-purpose oversampling methods, nor with pure cost-sensitive learning or algorithmic approaches. On top of oversampling, our method also assists the classifier in generalizing the minority class by considering information across both classes.

\newpage

\appendix

\section{Applied evaluation measures}
\label{appendix:measures}

For a classification of test data, let tp be the number of true positives, fn the number of false negatives, tn the number of true negatives, and fp the number of false positives.

The main evaluation measures applied in this article are
\begin{align*}
\text{balanced accuracy} &= \frac{\text{recall}+\text{tnr}}{2}, \\
\label{eq:fscore}
F_\beta &= (1+\beta^2) \frac{\text{precision} \times \text{recall}}{(\beta^2 \times \text{precision}) + \text{recall}},
\end{align*}
where recall, tnr, and precision are defined as
\begin{align*}
\text{recall} &= \frac{\text{tp}}{\text{tp} + \text{fn}}, \\
\text{tnr} &= \frac{\text{tn}}{\text{tn} + \text{fp}}, \\
\text{precision} &= \frac{\text{tp}}{\text{tp} + \text{fp}}.
\end{align*}

\section{Average recall, true negative rate, and precision}
\label{appendix:tables}

\begin{table}[ht]
\begin{center}
\scriptsize
\begin{tabular}{llllllllllll}
\\ \multicolumn{4}{l}{\scriptsize{\textbf{Very low frequency categories}}} & & & & & & & & \\
\multicolumn{2}{l}{\textbf{Sampling ratio: 10\% }} & SVM & ADASYN & DECOM & DRO & EDA & EMCO & EMCO & MCO & ROS & SMOTE \\
 & & & & & & & $\gamma$=1 & $\gamma$=0.1 & & & \\ \hline
\textbf{Reuters$^\dagger$} & \textbf{Titles} & .238 & .369 & .415$^3$ & .376 & .366 & {\bf .477}$^1$ & .472$^2$ & .360 & .368 & .369 \\
                             & \textbf{Full} & .312 & .426 & .538$^3$ & .462 & .441 & {\bf .604}$^1$ & .583$^2$ & .442 & .426 & .426 \\
\textbf{HuffPost}          & \textbf{Titles*} & .031 & .219 & .235 & .219 & .237$^3$ & {\bf .287}$^1$ & .271$^2$ & .208 & .207 & .217 \\
                             & \textbf{Titles} & .066 & .414$^3$ & .397 & NA & .423$^2$ & .413 & {\bf .429}$^1$ & .381 & .389 & .409 \\
                             & \textbf{Full*} & .033 & .184 & .243$^3$ & .200 & .209 & {\bf .336}$^1$ & .324$^2$ & .181 & .184 & .184 \\
\multicolumn{2}{l}{\textbf{Sampling ratio: 20\% }} & & & & & & & & & & \\ \hline
\textbf{Reuters$^\dagger$} & \textbf{Titles} & .238 & .368 & .455$^3$ & .377 & .359 & {\bf .511}$^1$ & .499$^2$ & .364 & .368 & .368 \\
                             & \textbf{Full} & .312 & .426 & .553$^3$ & .467 & .435 & {\bf .635}$^1$ & .621$^2$ & .461 & .426 & .426 \\
\textbf{HuffPost}          & \textbf{Titles*} & .031 & .231 & .352$^3$ & .225 & .260 & {\bf .395}$^1$ & .359$^2$ & .241 & .210 & .231 \\
                             & \textbf{Titles} & .066 & .440 & .502$^3$ & NA & .467 & {\bf .578}$^1$ & .572$^2$ & .439 & .403 & .440 \\
                             & \textbf{Full*} & .033 & .184 & .316$^3$ & .205 & .230 & {\bf .468}$^1$ & .437$^2$ & .216 & .184 & .184 \\
\multicolumn{4}{l}{\scriptsize{\textbf{Low frequency categories}}} & & & & & & & & \\
\multicolumn{2}{l}{\textbf{Sampling ratio: 10\% }} & & & & & & & & & & \\ \hline
\textbf{Reuters$^\dagger$} & \textbf{Titles} & .582 & .691 & .700 & .697 & .707$^2$ & {\bf .715}$^1$ & .701$^3$ & .680 & .684 & .680 \\
                             & \textbf{Full} & .720 & .800 & .807$^3$ & .802 & .806 & .839$^2$ & {\bf .846}$^1$ & .801 & .785 & .788 \\
\textbf{HuffPost}          & \textbf{Titles*} & .225 & .430$^3$ & .399 & .437$^2$ & {\bf .457}$^1$ & .399 & .411 & .395 & .425 & .426 \\
                             & \textbf{Titles} & .305 & .543$^3$ & .529 & NA & {\bf .577}$^1$ & .463 & .501 & .528 & .556$^2$ & .539 \\
                             & \textbf{Full*} & .245 & .439 & .430 & .446$^3$ & {\bf .464}$^1$ & .420 & .449$^2$ & .397 & .430 & .428 \\
\multicolumn{2}{r}{\textbf{20Newsgroups$^\dagger$}} & .477 & .556 & .560$^3$ & {\bf .567}$^1$ & .520 & .546 & .563$^2$ & .547 & .549 & .550 \\
\multicolumn{2}{l}{\textbf{Sampling ratio: 20\% }} & & & & & & & & & & \\ \hline
\textbf{Reuters$^\dagger$} & \textbf{Titles} & .582 & .716 & .778$^2$ & .726 & .742 & {\bf .795}$^1$ & .774$^3$ & .737 & .713 & .705 \\
                             & \textbf{Full} & .720 & .804 & .834$^3$ & .821 & .819 & {\bf .918}$^1$ & .906$^2$ & .825 & .798 & .801 \\
\textbf{HuffPost}          & \textbf{Titles*} & .231 & .499 & .520$^3$ & .493 & .512 & {\bf .534}$^1$ & .529$^2$ & .471 & .476 & .494 \\
                             & \textbf{Titles} & .316 & .631 & .649$^2$ & NA & {\bf .659}$^1$ & .608 & .640$^3$ & .623 & .632 & .626 \\
                             & \textbf{Full*} & .258 & .474 & .535$^3$ & .496 & .512 & .584$^2$ & {\bf .590}$^1$ & .471 & .471 & .470 \\
\multicolumn{2}{r}{\textbf{20Newsgroups$^\dagger$}} & .477 & .599 & .631$^3$ & .610 & .549 & .653$^2$ & {\bf .661}$^1$ & .618 & .585 & .593 \\
\end{tabular}
\\
\scriptsize{$^\dagger$Averages of 5 repetitions per category. $^*$Averages of 5-fold split of the data set.}
\caption{Average recall.}
\label{tab:tpr}
\end{center}
\end{table}

\begin{table}[ht]
\begin{center}
\scriptsize
\begin{tabular}{llllllllllll}
\\ \multicolumn{4}{l}{\scriptsize{\textbf{Very low frequency categories}}} & & & & & & & & \\
\multicolumn{2}{l}{\textbf{Sampling ratio: 10\% }} & SVM & ADASYN & DECOM & DRO & EDA & EMCO & EMCO & MCO & ROS & SMOTE \\
 & & & & & & & $\gamma$=1 & $\gamma$=0.1 & & & \\ \hline
\textbf{Reuters$^\dagger$} & \textbf{Titles} & {\bf 1.000}$^1$ & .999 & .992 & .999 & .999 & .994 & .996 & .999 & .999$^3$ & .999$^2$ \\
                             & \textbf{Full} & {\bf 1.000}$^1$ & 1.000$^2$ & .998 & .999 & 1.000 & .998 & .998 & 1.000 & 1.000 & 1.000$^3$ \\
\textbf{HuffPost}          & \textbf{Titles*} & {\bf 1.000}$^1$ & .994 & .992 & .994 & .993 & .988 & .990 & .994$^2$ & .994$^3$ & .994 \\
                             & \textbf{Titles} & {\bf 1.000}$^1$ & .989 & .988 & NA & .988 & .988 & .988 & .991$^2$ & .989$^3$ & .989 \\
                             & \textbf{Full*} & {\bf 1.000}$^1$ & .997 & .994 & .997 & .997 & .991 & .992 & .998$^2$ & .997 & .997$^3$ \\
\multicolumn{2}{l}{\textbf{Sampling ratio: 20\% }} & & & & & & & & & & \\ \hline
\textbf{Reuters$^\dagger$} & \textbf{Titles} & {\bf 1.000}$^1$ & .999$^3$ & .978 & .999 & .998 & .990 & .992 & .998 & .999 & .999$^2$ \\
                             & \textbf{Full} & {\bf 1.000}$^1$ & 1.000$^2$ & .996 & 1.000 & .999 & .995 & .997 & .999 & 1.000$^3$ & 1.000 \\
\textbf{HuffPost}          & \textbf{Titles*} & {\bf 1.000}$^1$ & .993 & .970 & .993 & .991 & .970 & .976 & .991 & .994$^2$ & .993$^3$ \\
                             & \textbf{Titles} & {\bf 1.000}$^1$ & .984 & .971 & NA & .983 & .966 & .969 & .985$^3$ & .986$^2$ & .984 \\
                             & \textbf{Full*} & {\bf 1.000}$^1$ & .997$^3$ & .987 & .997 & .996 & .977 & .981 & .996 & .997$^2$ & .997 \\
\multicolumn{4}{l}{\scriptsize{\textbf{Low frequency categories}}} & & & & & & & & \\
\multicolumn{2}{l}{\textbf{Sampling ratio: 10\% }} & & & & & & & & & & \\ \hline
\textbf{Reuters$^\dagger$} & \textbf{Titles} & {\bf .997}$^1$ & .992 & .990 & .990 & .990 & .988 & .990 & .992 & .993$^3$ & .993$^2$ \\
                             & \textbf{Full} & {\bf .997}$^1$ & .995 & .995 & .995 & .995 & .992 & .993 & .995 & .995$^3$ & .995$^2$ \\
\textbf{HuffPost}          & \textbf{Titles*} & {\bf .998}$^1$ & .989 & .990$^2$ & .984 & .986 & .990 & .990 & .990$^3$ & .988 & .989 \\
                             & \textbf{Titles} & {\bf .997}$^1$ & .987 & .988 & NA & .985 & .990$^2$ & .990$^3$ & .988 & .986 & .987 \\
                             & \textbf{Full*} & {\bf .998}$^1$ & .990 & .991 & .989 & .989 & .992$^3$ & .991 & .993$^2$ & .991 & .991 \\
\multicolumn{2}{r}{\textbf{20Newsgroups$^\dagger$}} & {\bf .996}$^1$ & .993 & .994 & .991 & .992 & .994$^3$ & .993 & .994$^2$ & .993 & .993 \\
\multicolumn{2}{l}{\textbf{Sampling ratio: 20\% }} & & & & & & & & & & \\ \hline
\textbf{Reuters$^\dagger$} & \textbf{Titles} & {\bf .997}$^1$ & .988 & .977 & .987 & .986 & .971 & .980 & .987 & .988$^3$ & .990$^2$ \\
                             & \textbf{Full} & {\bf .997}$^1$ & .993 & .991 & .993 & .993 & .981 & .987 & .993 & .994$^3$ & .995$^2$ \\
\textbf{HuffPost}          & \textbf{Titles*} & {\bf .997}$^1$ & .976 & .972 & .974 & .974 & .969 & .972 & .979$^2$ & .977$^3$ & .977 \\
                             & \textbf{Titles} & {\bf .996}$^1$ & .968 & .970 & NA & .969 & .971 & .972$^3$ & .975$^2$ & .971 & .969 \\
                             & \textbf{Full*} & {\bf .997}$^1$ & .985 & .980 & .982 & .982 & .973 & .975 & .986$^2$ & .985 & .985$^3$ \\
\multicolumn{2}{r}{\textbf{20Newsgroups$^\dagger$}} & {\bf .996}$^1$ & .989 & .988 & .988 & .990$^2$ & .983 & .985 & .990 & .990$^3$ & .990 \\
\end{tabular}
\\
\scriptsize{$^\dagger$Averages of 5 repetitions per category. $^*$Averages of 5-fold split of the data set.}
\caption{Average true negative rate.}
\label{tab:tnr}
\end{center}
\end{table}

\begin{table}[ht]
\begin{center}
\scriptsize
\begin{tabular}{llllllllllll}
\\ \multicolumn{4}{l}{\scriptsize{\textbf{Very low frequency categories}}} & & & & & & & & \\
\multicolumn{2}{l}{\textbf{Sampling ratio: 10\% }} & SVM & ADASYN & DECOM & DRO & EDA & EMCO & EMCO & MCO & ROS & SMOTE \\
 & & & & & & & $\gamma$=1 & $\gamma$=0.1 & & & \\ \hline
\textbf{Reuters$^\dagger$} & \textbf{Titles} & .465 & .504$^3$ & .177 & .494 & .429 & .212 & .250 & .421 & .504$^2$ & {\bf .508}$^1$ \\
                             & \textbf{Full} & .549 & {\bf .633}$^1$ & .444 & .629 & .603 & .468 & .502 & .625 & .632$^3$ & .632$^2$ \\
\textbf{HuffPost}          & \textbf{Titles*} & {\bf .388}$^1$ & .239$^3$ & .187 & .228 & .222 & .167 & .180 & .232 & .232 & .240$^2$ \\
                             & \textbf{Titles} & {\bf .634}$^1$ & .243 & .218 & NA & .235 & .216 & .230 & .268$^2$ & .237 & .247$^3$ \\
                             & \textbf{Full*} & .373 & .396$^3$ & .246 & .372 & .365 & .233 & .251 & .388 & {\bf .397}$^1$ & .397$^2$ \\
\multicolumn{2}{l}{\textbf{Sampling ratio: 20\% }} & & & & & & & & & & \\ \hline
\textbf{Reuters$^\dagger$} & \textbf{Titles} & .465 & .503$^2$ & .077 & .494 & .370 & .148 & .179 & .391 & .502$^3$ & {\bf .503}$^1$ \\
                             & \textbf{Full} & .549 & .634$^2$ & .353 & {\bf .635}$^1$ & .584 & .369 & .409 & .600 & .632$^3$ & .632 \\
\textbf{HuffPost}          & \textbf{Titles*} & {\bf .388}$^1$ & .220 & .093 & .214 & .195 & .103 & .113 & .179 & .223$^2$ & .220$^3$ \\
                             & \textbf{Titles} & {\bf .634}$^1$ & .195 & .129 & NA & .192 & .129 & .141 & .196 & .204$^2$ & .197$^3$ \\
                             & \textbf{Full*} & .373 & .395$^2$ & .166 & .367 & .339 & .143 & .163 & .342 & {\bf .395}$^1$ & .395$^3$ \\
\multicolumn{4}{l}{\scriptsize{\textbf{Low frequency categories}}} & & & & & & & & \\
\multicolumn{2}{l}{\textbf{Sampling ratio: 10\% }} & & & & & & & & & & \\ \hline
\textbf{Reuters$^\dagger$} & \textbf{Titles} & {\bf .884}$^1$ & .707 & .656 & .685 & .685 & .616 & .670 & .726$^3$ & .725 & .746$^2$ \\
                             & \textbf{Full} & {\bf .886}$^1$ & .824 & .814 & .821 & .828 & .749 & .772 & .832 & .836$^3$ & .843$^2$ \\
\textbf{HuffPost}          & \textbf{Titles*} & {\bf .691}$^1$ & .489 & .492 & .424 & .457 & .486 & .493$^3$ & .501$^2$ & .474 & .492 \\
                             & \textbf{Titles} & {\bf .724}$^1$ & .504 & .525 & NA & .499 & .521 & .533$^3$ & .535$^2$ & .505 & .512 \\
                             & \textbf{Full*} & {\bf .740}$^1$ & .534 & .533 & .505 & .519 & .547 & .534 & .573$^2$ & .536 & .547$^3$ \\
\multicolumn{2}{r}{\textbf{20Newsgroups$^\dagger$}} & {\bf .860}$^1$ & .803 & .807$^3$ & .767 & .770 & .806 & .799 & .814$^2$ & .802 & .805 \\
\multicolumn{2}{l}{\textbf{Sampling ratio: 20\% }} & & & & & & & & & & \\ \hline
\textbf{Reuters$^\dagger$} & \textbf{Titles} & {\bf .884}$^1$ & .640 & .491 & .624 & .607 & .438 & .518 & .626 & .642$^3$ & .687$^2$ \\
                             & \textbf{Full} & {\bf .886}$^1$ & .789 & .724 & .790 & .786 & .580 & .659 & .769 & .807$^3$ & .833$^2$ \\
\textbf{HuffPost}          & \textbf{Titles*} & {\bf .695}$^1$ & .413 & .363 & .388 & .394 & .350 & .372 & .420$^2$ & .410 & .416$^3$ \\
                             & \textbf{Titles} & {\bf .727}$^1$ & .391 & .408 & NA & .417 & .389 & .417 & .447$^2$ & .425$^3$ & .399 \\
                             & \textbf{Full*} & {\bf .739}$^1$ & .508 & .445 & .487 & .488 & .403 & .424 & .521$^2$ & .508 & .515$^3$ \\
\multicolumn{2}{r}{\textbf{20Newsgroups$^\dagger$}} & {\bf .860}$^1$ & .752 & .735 & .731 & .738 & .677 & .704 & .759$^2$ & .754 & .756$^3$ \\
\end{tabular}
\\
\scriptsize{$^\dagger$Averages of 5 repetitions per category. $^*$Averages of 5-fold split of the data set.}
\caption{Average precision.}
\label{tab:prec}
\end{center}
\end{table}

\clearpage

\bibliographystyle{apalike}
\bibliography{references}

\end{document}